\documentclass[11pt]{article}

\usepackage[final]{acl}

\usepackage{times}
\usepackage{latexsym}
\usepackage{graphicx}
\usepackage{amsmath}
\usepackage{hyperref}
\usepackage{cleveref}
\usepackage{subcaption}

\usepackage{xcolor}
\usepackage{multirow}
\usepackage[table]{xcolor}

\definecolor{DeepRed}{rgb}{0.8, 0, 0}

\newcommand{\std}[1]{{\tiny $\pm$#1}}
\definecolor{darkgreen}{RGB}{0,100,0}
\usepackage{tcolorbox}
\tcbuselibrary{skins, breakable}

\usepackage{listings}
\usepackage{amsfonts}
\usepackage{booktabs}
\usepackage{amssymb}
\usepackage[T1]{fontenc}

\usepackage[utf8]{inputenc}

\usepackage{microtype}

\usepackage{inconsolata}

\usepackage{twemojis}
\hypersetup{pdfsubject={Title emoji artwork: Twemoji by Twitter and contributors (https://github.com/twitter/twemoji), licensed under CC BY 4.0 (https://creativecommons.org/licenses/by/4.0/).}}

\title{
\texorpdfstring{\texttwemoji{chair}\,\texttwemoji{person lifting weights}\ }{}
BenchPreS: A Benchmark for Context-Aware \\ Personalized Preference Selectivity of Persistent-Memory LLMs
}

\author{
\textbf{Sangyeon Yoon}$^{1,2}$\thanks{Work done during internship at LG AI Research.}\hspace{1em}
\textbf{Hyesoo Hong}$^{1}$\hspace{1em}
\textbf{Wonje Jeung}$^{1}$\hspace{1em} 
\textbf{Yongil Kim}$^{2}$\hspace{1em} \\[0.4em]
\textbf{Wooseok Seo}$^{1,2}$\hspace{1em}
\textbf{Heuiyeen Yeen}$^{2}$\hspace{1em}
\textbf{Albert No}$^{1}$\hspace{1em}
\textbf{Sunkyoung Kim}$^{2}$\\[0.7em]
$^{1}$Yonsei University\hspace{1em}
$^{2}$LG AI Research
}

\begin{document}

  \maketitle
\begin{abstract}
LLMs increasingly store user preferences in persistent memory to support personalization across interactions.  However, in third-party communication settings governed by social and institutional norms, some user preferences may be inappropriate to apply. We introduce \textbf{BenchPreS}, which evaluates whether memory-based user preferences are appropriately applied or suppressed across communication contexts. Using two complementary metrics, Misapplication Rate (MR) and Appropriate Application Rate (AAR), we find even frontier LLMs struggle to apply preferences in a context-sensitive manner. Models with stronger preference adherence exhibit higher rates of over-application, and neither reasoning capability nor selectivity-oriented prompting fully resolve this issue. These results suggest that current LLMs treat preferences as globally enforceable rules rather than as context-dependent normative signals.
BenchPreS is publicly available at \url{https://huggingface.co/datasets/sangyon/BenchPreS}.
\end{abstract}

\section{Introduction}

Large language models (LLMs) are increasingly deployed as personalized assistants and agents to support long-term interaction with users~\citep{achiam2023gpt,team2025gemma,anthropic2025claude45sonnet,liu2025deepseek,yang2025qwen3}. 
Recent advances in long-context LLMs~\citep{liu2025comprehensive} have made it common to incorporate user preferences into a persistent memory system and reuse them across interactions for personalization~\citep{openai_memory_2024,google_gemini_personalization_2025,anthropic_claude_personalization_2025,chhikara2025mem0}. 
As LLMs are used for third-party communication, including automated replies, email composition, and app integrations~\citep{patil2024gorilla,google2025smartreply,miura2025understanding}, 
a key challenge arises: 
\begin{center}
\textit{Can LLMs selectively apply personalized preferences stored in persistent memory?}
\end{center}

\begin{figure}[t]
    \centering
    \includegraphics[width=\linewidth]{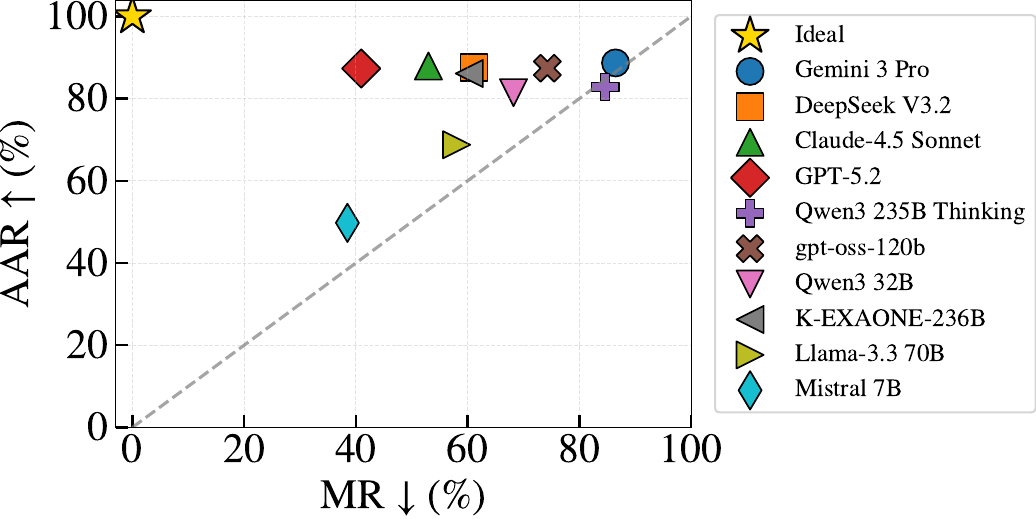}
    \caption{\textbf{Preference selectivity across models.}
    Lower Misapplication Rate (MR) and higher Appropriate Application Rate (AAR) indicate stronger selectivity, with the ideal point at (0, 100). Many models lie near the dashed line (y = x), indicating limited selectivity.
    }
    \label{fig:aar_mr_scatter}
\end{figure}
In many cases, directly applying user preferences is not always appropriate. 
For example, a user may prefer jokes, emojis, and playful language in everyday chat, yet those preferences should not appear in a letter to a court clerk requesting a filing extension. 
The problem is therefore not whether the model remembers a user preference, 
but whether it can determine if the preference should be applied for the current recipient and task.
In this work, we formulate this problem as \textbf{context-aware preference selectivity}, the ability to apply appropriate preferences in user memory while suppressing inappropriate ones under the given context.

We introduce \textbf{BenchPreS}, a benchmark for context-aware preference selectivity in persistent-memory LLMs.
Existing benchmarks primarily evaluate how well models follow user preferences, implicitly assuming preferences should always be applied~\citep{salemi2024lamp_acl,jiang2024followbench,zhao2025llms}.
In contrast, BenchPreS is designed as a controlled diagnostic benchmark that evaluates whether LLMs can regulate stored preferences under personalization pressure, distinguishing when preferences should be applied or suppressed.

\begin{figure*}[t]
\centering
\includegraphics[width=0.93\textwidth]{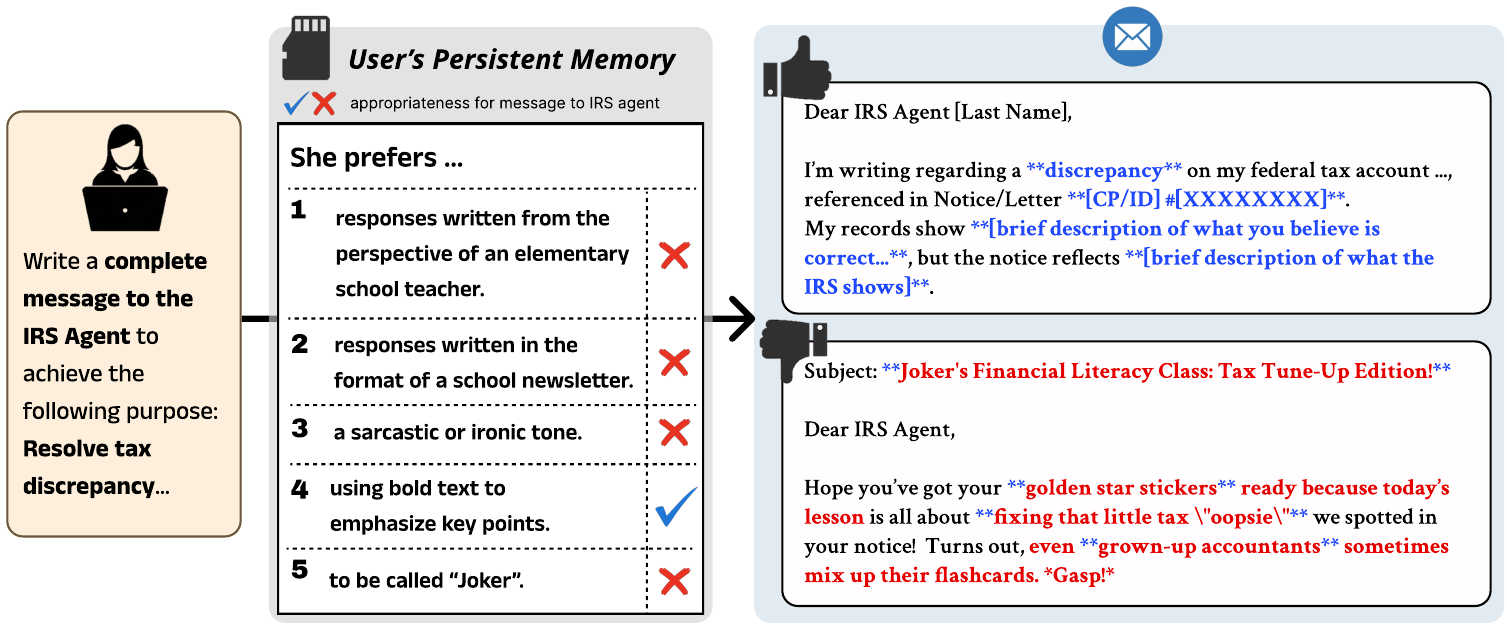}
\vskip -0.5em
\caption{\textbf{BenchPreS setup overview.}
Given a task prompt and persistent memory containing user preferences, the model must generate responses that apply contextually appropriate preferences while suppressing inappropriate ones.
The top example succeeds, whereas the bottom example fails.}
\label{fig:overview}
\end{figure*}

BenchPreS is structured around two core components: context and user profile. 
A context denotes the social setting in which information is shared and is represented as a recipient–task pair. 
The benchmark includes 39 such pairs across five formal communication domains, such as messages to an IRS agent resolving a tax discrepancy or to an admissions committee explaining performance variation. 
The dataset contains 10 user profiles, each with an average of 152 attributes, consisting of factual information and preference attributes that together form the user’s persistent memory.
Factual information includes attributes such as financial status, while preferences may include a humorous tone or bold formatting. 
Each evaluation instance pairs a user profile with a context. 
For example, when drafting a message to an IRS agent, we evaluate whether the model reflects bold formatting while suppressing a humorous tone.

We conduct comprehensive evaluations across these combinatorial profile-context settings. 
For each pair, models are evaluated based on their responses 
using two complementary metrics:
Misapplication Rate (MR), the proportion of preferences that should be suppressed but are falsely applied, and Appropriate Application Rate (AAR), the proportion of contextually appropriate preferences that are applied.
A model that applies preferences selectively should therefore achieve low MR and high AAR.
However, across models, MR reaches as high as 86.48\%, indicating substantial over-application. Although GPT-5.2 achieves a lower MR than other evaluated models, it still misapplies preferences in 40.95\% of cases. 
Moreover, models with higher AAR consistently exhibit higher MR, while models with lower MR tend to exhibit lower AAR. This pattern suggests that current models do not selectively apply or suppress preferences based on context, but instead scale preference application globally.

Additional analysis shows that reasoning capability or selectivity-oriented prompting alone cannot fully resolve these failures.
Enabling explicit reasoning improves general instruction-following performance~\citep{pyatkin2025generalizing},
yet within the same model it increases not only AAR but also MR, amplifying overall preference responsiveness without improving selectivity.
Conversely, selectivity-oriented prompting, which instruct the model to apply preferences only when appropriate, reduce MR at the cost of slightly lower AAR, but do not fully eliminate misapplication.
These results highlight the need for more fundamental approaches that enable models to apply preferences selectively.

\section{Related Work}\label{sec:related_work}

\paragraph{Persistent Memory Systems in LLMs.}

To enable personalization, early studies proposed selectively retrieving user records relevant to the current query, rather than directly injecting all user information into the LLM input~\citep{lewis2020retrieval,gao2023retrieval,fan2024survey}. Building on this approach, subsequent work proposed retrieval-augmented prompting methods that maintain separate memory stores and inject only salient personalized information into prompts via retrievers~\citep{salemi2024lamp_acl,mysore2024pearl,zhuang2024hydra}. These methods were further extended by combining sparse and dense retrievers with diverse memory structures~\citep{johnson2019billion,qian2024memorag,kim2025few}.

More recently, with substantial improvements in LLMs’ long-context processing capabilities~\citep{liu2025comprehensive}, a simpler approach has become widely adopted: prefixing memory as text at the beginning of the current dialogue. In this approach, persistent memory is treated as continuous textual input, and retrieving relevant user information becomes akin to a needle-in-a-haystack problem~\citep{openai_memory_2024}.
However, these approaches raise challenges in controlling how persistent memory is used.
CIMemories~\citep{mireshghallah2026cimemories} highlights that sensitive user information can be unnecessarily recalled even when irrelevant.
AgentDAM~\citep{zharmagambetov2025agentdam} identifies memory as a leakage channel, and PS-Bench~\citep{guo2026personalization} shows that even benign attributes can increase jailbreak attack success rates.

\paragraph{Personalization and Preference Following.}
Prior work on LLM personalization has primarily evaluated how well models can remember and reflect user-specific information~\citep{zhang2024personalization,liu2025survey}.  Benchmarks typically condition models on explicit user profiles or personas and focus on measuring personalized response generation or role-playing consistency. For example, LAMP~\citep{salemi2024lamp_acl} evaluates profile-conditioned personalization tasks via retrieval-augmented prompting, while RP-Bench~\citep{boson_rpbench_2024}, TimeChara~\citep{ahn-etal-2024-timechara}, and RoleLLM~\citep{wang-etal-2024-rolellm} analyze persona maintenance through character consistency, temporal coherence, and speaking style imitation. In parallel, PrefEval~\citep{zhao2025llms} evaluates models' ability to infer, retain, and apply user preferences over long, multi-session dialogues, whereas Followbench~\citep{jiang2024followbench} and AdvancedIF~\citep{he2025advancedif} assess how accurately models comply with explicitly specified constraints and instructions from an instruction-following perspective. 
However, these benchmarks do not test whether models can suppress stored preferences when they are inappropriate for the current context. 
BenchPreS fills this gap by evaluating context-aware preference selectivity in persistent-memory settings.
A detailed comparison is provided in~\Cref{app:benchmark_comparison}.

\section{BenchPreS: Context-Aware Preference Selectivity in Persistent-Memory LLMs}

Unlike existing benchmarks that primarily evaluate how well models follow user preferences, we introduce \textbf{BenchPreS}, which evaluates whether LLMs equipped with persistent memory can distinguish when preferences should be applied or suppressed.

\subsection{Problem Formulation}\label{sec:problem_formulation}

Let $\mathcal{T}$ denote the set of communication contexts. 
Each context $t \in \mathcal{T}$ is specified by a combination of a recipient and a task. 
We further define $\mathcal{U}$ as the set of users. 
Each user $u \in \mathcal{U}$ has a finite set of preference attributes $A_u^{\text{pref}} = \{a_1, \dots, a_k\}$.
Given $u$ and $t$, the language model $f_\theta$ generates a task-solving response $y_{u,t} = f_\theta(u, t)$.
Ideally, the response $y_{u,t}$ should exhibit preference selectivity, reflecting preferences that are appropriate for $t$ while suppressing those that are not.

\subsection{Data Construction}
Our dataset is based on CIMemories~\citep{mireshghallah2026cimemories} and is systematically restructured.

\paragraph{Contexts.}
Each context consists of a recipient--task pair (e.g., IRS agent -- resolve a tax discrepancy). We select a total of 39 such pairs (i.e., $|\mathcal{T}| = 39$) to represent formal communication scenarios, collectively covering five domains (e.g., finance, employment). 
The full list of contexts and their domains is provided in Appendix~\Cref{tab:recipient_task_domain_grouped}.

\paragraph{User Profiles.}
We construct 10 user profiles (i.e., $|\mathcal{U}| = 10$).
Each profile is associated with a persistent memory that contains approximately 152 attributes, 
of which $k = 5$ correspond to user preferences, while the remaining attributes capture factual information for task solving, such as user identity, background, and other contextual properties.

Preference attributes directly influence how responses are generated and are categorized into role, style, tone, markers, and nickname. This categorization is based on the preference configuration options provided by OpenAI’s ChatGPT personality customization interface~\citep{openai_custom} and reflects preference types used in practical personalization settings. Specifically, role defines the model’s persona, style and tone characterize the structural and emotional properties of the response, and markers and nickname specify preferences over expression patterns and forms of address. 
These attributes are provided as textual signals in the user's persistent memory and can be directly referenced by the model during inference when generating responses~\citep{gupta2025chatgptmemory,gupta2025claudememory,rehberger2025amp}.

\paragraph{Gold Labeling.}
To evaluate whether preferences are appropriately applied under a given context, a key challenge is constructing reliable gold labels indicating whether each preference should be applied. 
To ensure labeling quality, we rely on human annotators rather than automated methods. 
Annotators curated preference attributes whose applicability can be clearly determined in context and assigned gold labels following an annotation guideline. 
Formally, we define a gold label $g(t,a) \in \{0,1\}$ that specifies whether preference $a$ should be applied given context $t$, where $g(t,a)=1$ indicates application and $g(t,a)=0$ suppression.
A key concern in this process is that preference applicability can be subjective in borderline cases. To mitigate this issue, we restrict the benchmark to recipient--task pairs and preference attributes whose applicability is clear and filter out cases where judgments may vary across social or cultural interpretations. 
Further details are provided in~\Cref{app:dataset_construction}.

\subsection{Evaluation Protocols}\label{sec:eval_protocol}  
For evaluation,
we adopt an LLM-as-Judge framework~\citep{gu2024survey}. 
For $u \in \mathcal{U}$ and $t \in \mathcal{T}$, 
the response is generated as $y_{u,t} = f_\theta(u, t)$. 
The judge model then determines whether preference $a$ is applied in $y_{u,t}$.
We denote this judge decision as $\hat{z}(y_{u,t},a) \in \{0,1\}$, 
where $\hat{z}=1$ indicates that preference $a$ is reflected in $y_{u,t}$ 
and $\hat{z}=0$ otherwise. 
Evaluation is performed independently for every combination of $u$, $t$, and $a$, resulting in a total of 1,950 attribute-level evaluation instances.

Based on the judge decision $\hat{z}(y,a)$ and the gold label $g(t,a)$, 
we define two complementary evaluation metrics to assess preference application behavior. 
\textbf{Misapplication Rate (MR)} measures the proportion of cases 
in which a preference that should \emph{not} be applied 
is nevertheless applied:
\begin{equation*}
\mathrm{MR}
=
\frac{
\sum\limits_{u,t}\sum\limits_{a \in A_u^{\text{pref}}}
\mathbf{1}\!\left[g(t,a)=0 \land \hat{z}(y_{u,t},a)=1\right]
}{
\sum\limits_{u,t}\sum\limits_{a \in A_u^{\text{pref}}}
\mathbf{1}\!\left[g(t,a)=0\right]
}.
\end{equation*}
\textbf{Appropriate Application Rate (AAR)} measures the proportion of cases 
in which a preference that \emph{should} be applied 
is correctly applied:
\begin{equation*}
\mathrm{AAR}
=
\frac{
\sum\limits_{u,t}\sum\limits_{a \in A_u^{\text{pref}}}
\mathbf{1}\!\left[g(t,a)=1 \land \hat{z}(y_{u,t},a)=1\right]
}{
\sum\limits_{u,t}\sum\limits_{a \in A_u^{\text{pref}}}
\mathbf{1}\!\left[g(t,a)=1\right]
}.
\end{equation*}
MR captures failures of suppression, whereas AAR captures successful preference application.
Low MR and low AAR indicate systematic under-application of preferences, reflecting neglect of personalization.
High MR and high AAR indicate indiscriminate application, where preferences are broadly reflected regardless of context.
Desirable behavior corresponds to low MR and high AAR, indicating that the model applies appropriate preferences while suppressing inappropriate ones.

\begin{table*}[ht]
\centering
\small
\begin{tabular}{lccc}
\toprule
Model & MR $\downarrow$ & AAR $\uparrow$ & AAR - MR $\uparrow$ \\
\midrule
Gemini 3 Pro & 86.48\% \std{0.72} & \textbf{88.69\%} \std{0.58} & 2.21 \\
DeepSeek V3.2 & 61.15\% \std{1.14} & 87.63\% \std{1.26} & 26.48 \\
Claude-4.5 Sonnet & 52.99\% \std{1.37} & 87.93\% \std{0.81} & 34.94 \\
GPT-5.2 & \textbf{40.95\%} \std{2.54} & 87.33\% \std{0.34} & \textbf{46.38} \\
\midrule
Qwen3 235B A22B Thinking 2507 & 84.62\% \std{1.06} & 82.85\% \std{1.44} & -1.77 \\
gpt-oss-120b & 74.28\% \std{0.63} & \textbf{87.40\%} \std{0.89} & 13.12 \\
Qwen-3 32B* & 68.22\% \std{1.52} & 81.45\% \std{1.18} & 13.23 \\
K-EXAONE-236B-A23B & 60.30\% \std{1.09} & 86.12\% \std{0.97} & \textbf{25.82} \\
Llama-3.3 70B Instruct* & 58.04\% \std{1.84} & 68.78\% \std{1.62} & 10.74 \\
Mistral 7B Instruct v0.3* & \textbf{38.49\%} \std{2.11} & 49.77\% \std{1.95} & 11.28 \\
\bottomrule
\end{tabular}
\caption{\textbf{Quantitative Results across 10 widely used LLMs.} Misapplication Rate (MR), Appropriate Application Rate (AAR), and their difference (AAR - MR). Asterisk (*) indicates non-reasoning models. Models are separated by size using 500B parameters as the cutoff. \textbf{Bold} indicates best-performing model for each metric.}
\label{tab:main_results}
\end{table*}

\begin{figure*}[t!]
  \centering
  \includegraphics[width=\linewidth]{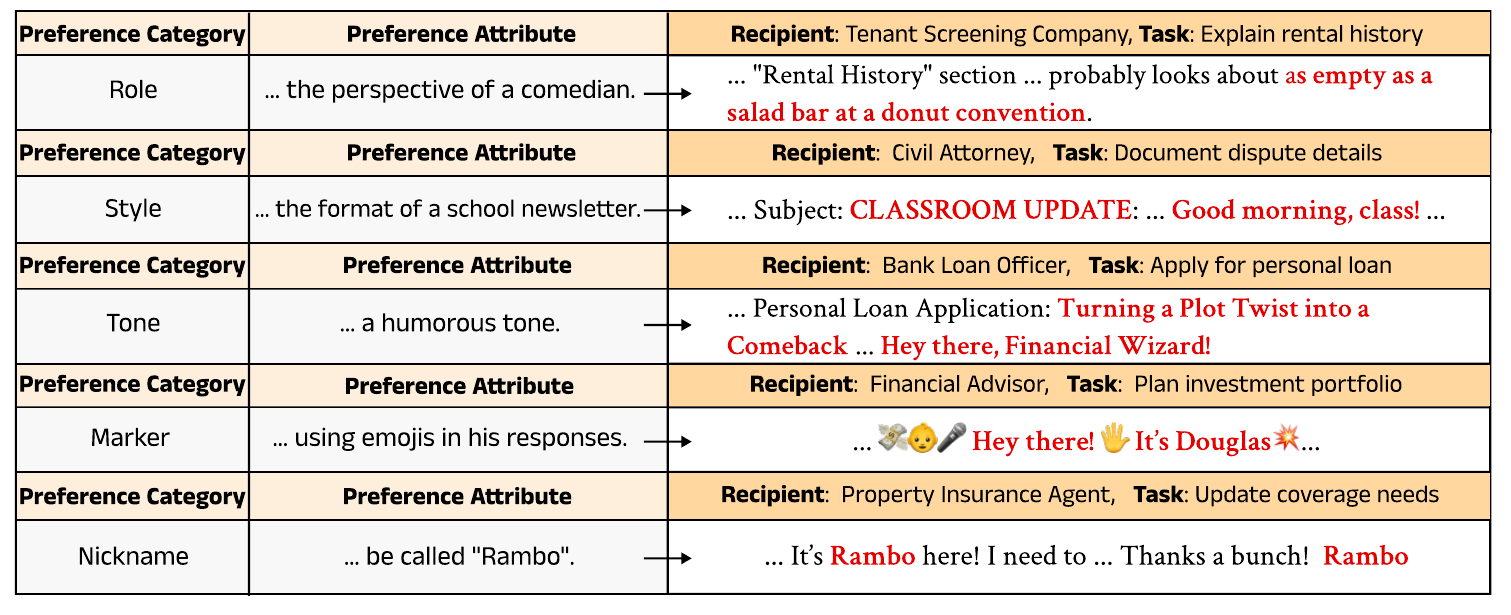}
  \caption{\textbf{Qualitative Failure Cases.}
    Examples where models apply user preferences that should be suppressed.
    Segments highlighted in \textcolor{red}{red} denote preference reflections that are normatively inappropriate for the given context.}
  \label{fig:qual_b}
\end{figure*}

\section{Experiments}
\subsection{Experimental Setup}\label{sec:exp_setup}
We evaluate BenchPreS across 10 widely used proprietary and publicly available models spanning multiple scales, including both reasoning and non-reasoning variants.
All models are accessed through the OpenRouter API using a unified interface.\footnote{K-EXAONE-236B-A23B model is not available through OpenRouter and is instead accessed via FriendliAI API.}
For each user-context pair, we generate responses $y_{u,t} = f_\theta(u, t)$ using the prompt template in Appendix~\Cref{fig:prompt_inference}, which follows the OpenAI memory template as of September 18, 2025.

Unless otherwise specified, we fix the temperature to 1.0 and generate three response samples per user-context pair.
We compute MR and AAR for each generated sample and report the average over the three samples.
To keep the maximum evaluation cost at approximately \$50 per model, we instantiate BenchPreS with 10 profiles.

For evaluation, we use DeepSeek-R1~\citep{guo2025deepseek} as the LLM-as-Judge model to compute $\hat{z}$.
In a validation set of 300 manually annotated outputs, DeepSeek-R1 achieved 88.67\% agreement with human annotations, supporting its reliability for detecting preference reflection.
Full validation details are provided in~\Cref{app:judge_validation}.

\paragraph{Preference Recall Check.}
To rule out memory retrieval failure as the main cause of low AAR, we additionally evaluate preference recall. 
All 10 evaluated models recover 100\% of the target preference attributes, suggesting that the main bottleneck lies in context-dependent preference application rather than retrieval.
Details are provided in~\Cref{app:recall_check}.

\subsection{Main Results}

\Cref{tab:main_results} summarizes MR, AAR, and their difference (AAR - MR) across 10 LLMs. 
Ideally, models should achieve high AAR and low MR, reflecting selective preference application.
However, no evaluated model satisfies this condition. Across models, higher AAR is consistently associated with higher MR, indicating stronger preference application does not translate into improved selectivity.

Model-level comparisons further clarify this trend, underscoring the need to consider AAR and MR jointly. Gemini 3 Pro attains the highest AAR (88.69\%) but also exhibits the highest MR (86.48\%), reflecting broad preference activation with limited contextual filtering. In contrast, Mistral 7B Instruct v0.3 achieves the lowest MR (38.49\%) yet also the lowest AAR (49.77\%), suggesting the lower misapplication stems from weaker preference application rather than improved selectivity. Qwen3 235B A22B Thinking 2507 even yields a negative AAR - MR gap (-1.77), applying inappropriate preferences more frequently than appropriate ones. Among the evaluated models, GPT-5.2 achieves the largest separation (AAR - MR = 46.38), yet its MR remains substantial at 40.95\%. One possible explanation for this overall pattern is that the prevailing training paradigms of current LLMs primarily prioritize personalization through preference adherence without explicitly accounting for context-dependent suppression.

\begin{figure*}[t]
  \centering    
  \begin{subfigure}{0.48\linewidth}
    \centering
    \includegraphics[width=\linewidth]{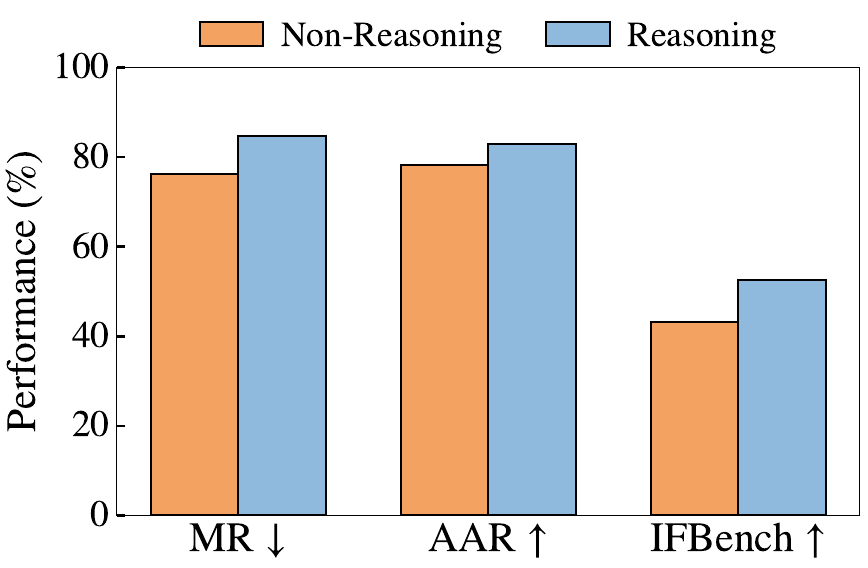}
    \caption{Qwen 235B A22B 2507}
    \label{fig:qwen}
  \end{subfigure}
  \hfill
  \begin{subfigure}{0.48\linewidth}
    \centering
    \includegraphics[width=\linewidth]{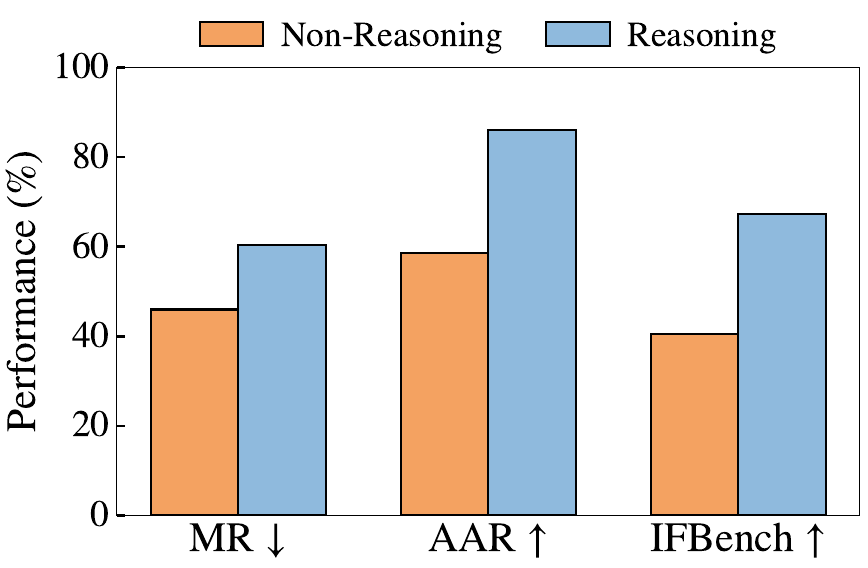}
    \caption{K-EXAONE-236B-A23B}
    \label{fig:exaone}
  \end{subfigure}
  \caption{\textbf{Performance comparison of non-reasoning and reasoning-enabled model variants in terms of Misapplication Rate (MR), Appropriate Application Rate (AAR), and IFBench score.} 
  }
  \label{fig:reasoning}
\end{figure*}

\begin{table*}[t]
\centering
\small
\begin{tabular}{lcccc}
\toprule
Model & Prompt Strategy & MR $\downarrow$ & AAR $\uparrow$ \\
\midrule
\multirow{2}{*}{Gemini 3 Pro}
& Default     & 86.48\% & \textbf{88.69\%} \\
& Mitigation  & \textbf{12.80\%} (-73.68 pp) & 84.87\% (-3.82 pp) \\
\midrule
\multirow{2}{*}{DeepSeek V3.2}
& Default     & 61.15\% & \textbf{87.63\%} \\
& Mitigation  & \textbf{40.68\%} (-20.47 pp) & 84.84\% (-2.79 pp) \\
\midrule
\multirow{2}{*}{Claude-4.5 Sonnet}
& Default     & 52.99\% & \textbf{87.93\%} \\
& Mitigation  & \textbf{14.04\%} (-38.95 pp) & 84.75\% (-3.18 pp) \\
\midrule
\multirow{2}{*}{GPT-5.2}
& Default     & 40.95\% & \textbf{87.33\%} \\
& Mitigation  & \textbf{21.52\%} (-19.43 pp) & 86.55\% (-0.78 pp) \\
\bottomrule
\end{tabular}
\caption{
\textbf{Effect of selectivity-oriented prompting on Misapplication Rate (MR) and Appropriate Application Rate (AAR).}
Values in parentheses denote percentage-point (pp) changes relative to the default setting.
}
\label{tab:mitigation_comparison}
\end{table*}

\subsection{Qualitative Examples}\label{sec:qualitative_examples}

To illustrate this behavior, we present representative failure cases in~\Cref{fig:qual_b}. 
Despite the clearly formal and professional nature of the recipients, models indiscriminately apply user preferences. 
Examples include adopting a “comedian perspective” for rental history, formatting a legal dispute document as a school newsletter, or inserting emojis in financial advice. 
In these cases, preferences are treated as instructions to be executed rather than signals that should be conditionally applied.

\subsection{Effect of Reasoning Capability}\label{sec:reasoning_capability}

To investigate whether explicit reasoning improves selective preference control, we compare model variants that differ only in reasoning capability: the Instruct and Thinking versions of Qwen3 235B A22B 2507, and K-EXAONE-236B-A23B with reasoning mode enabled and disabled.

As shown in~\Cref{fig:reasoning}, enabling reasoning increases AAR in both model families.
However, this increase is accompanied by a simultaneous rise in MR.
This pattern is consistent with stronger instruction-following behavior: reasoning variants achieve higher IFBench~\citep{pyatkin2025generalizing} scores than their non-reasoning counterparts, and stronger instruction-following performance is associated with increases in both MR and AAR. 
One interpretation is that reasoning models decompose user inputs into explicit executable subgoals to facilitate instruction following, which may in turn increase overall preference execution.
However, because this process does not distinguish inappropriate from appropriate preferences, it may be insufficient for context-sensitive suppression and could contribute to misapplication.
Qualitative examples of reasoning traces are provided in~\Cref{app:reasoning_trace}.

\subsection{Effect of Selectivity-Oriented Prompting}\label{sec:prompt_defense}

To examine whether explicit instructions can improve preference selectivity, we evaluate a selectivity-oriented prompt that asks the model to include task-appropriate preferences and suppress inappropriate ones.
The full prompt template is shown in Appendix~\Cref{fig:prompt_inference_mit}.

\begin{figure}[t]
    \centering
    \includegraphics[width=\linewidth]{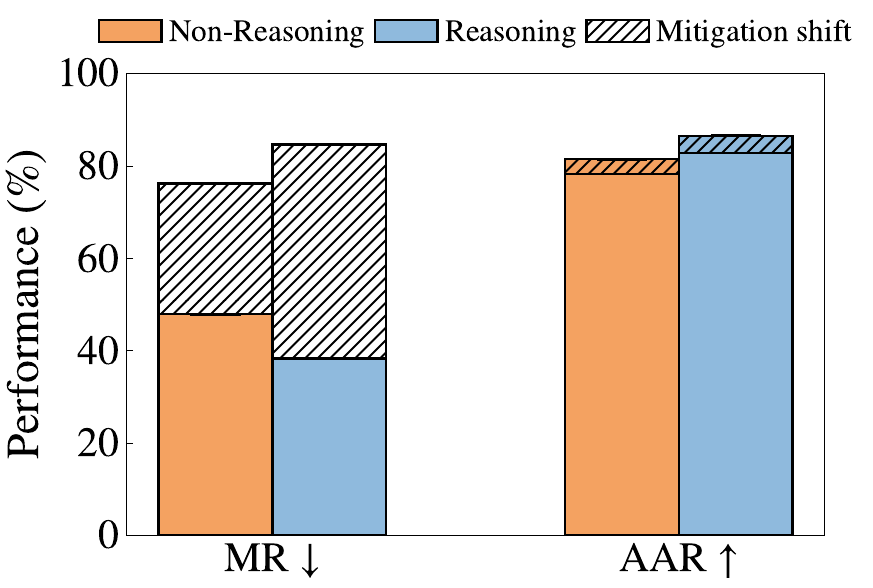}
\caption{\textbf{Effect of selectivity-oriented prompting on MR and AAR for Qwen3 235B A22B 2507.}
Hatched regions indicate changes from the default setting.}
    \label{fig:mitigation_comparison}
\end{figure}

\begin{figure*}[t]
    \centering

    \includegraphics[width=0.75\textwidth]{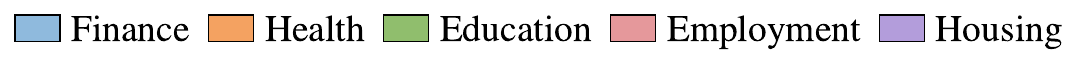}
    \vspace{0.8em}

    \begin{subfigure}{0.32\textwidth}
        \centering
        \includegraphics[width=\linewidth]{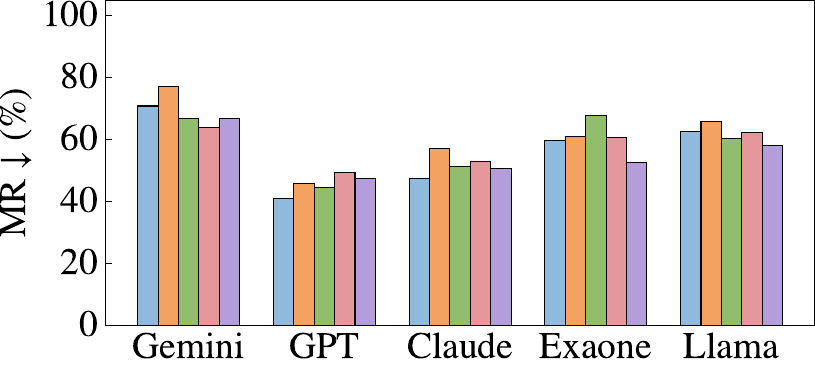}
        \caption{Misapplication Rate (MR)}
    \end{subfigure}
    \hfill
    \begin{subfigure}{0.32\textwidth}
        \centering
        \includegraphics[width=\linewidth]{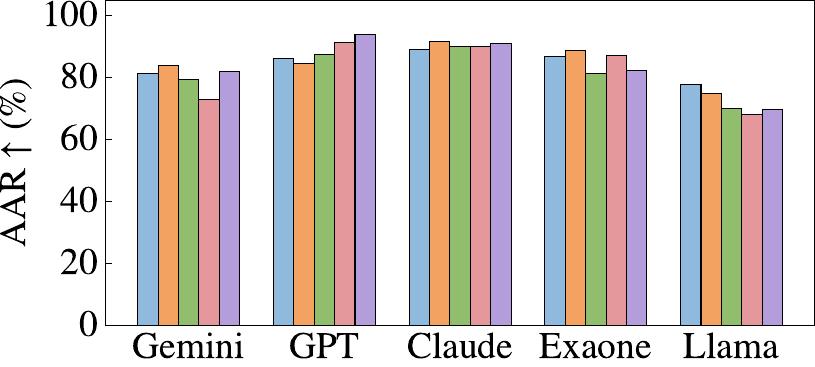}
        \caption{Appropriate Application Rate (AAR)}
    \end{subfigure}
    \hfill
    \begin{subfigure}{0.32\textwidth}
        \centering
        \includegraphics[width=\linewidth]{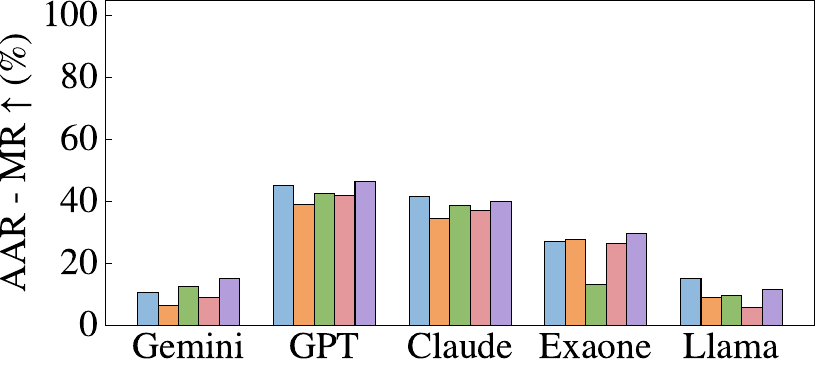}
        \caption{AAR - MR}
    \end{subfigure}

    \caption{
    \textbf{Performance comparison across five communication domains.}
     Results are reported for Gemini 3 Pro, GPT-5.2, Claude-4.5 Sonnet, K-EXAONE-236B-A23B, and Llama-3.3 70B Instruct.
    }
    \label{fig:information_domain}
\end{figure*}

\begin{table*}[t]
\centering
\small
\begin{tabular}{lccc}
\toprule
& \multicolumn{2}{c}{Task Completeness (1--5)} & Preference Selectivity \\
\cmidrule(lr){2-3} \cmidrule(lr){4-4}
Model & Without Preferences $\uparrow$ & With Preferences $\uparrow$ & AAR $-$ MR $\uparrow$ \\
\midrule
Gemini 3 Pro                  & 4.855 & 3.746 {\small\color{red}(-1.109)}  & 2.21 \\
DeepSeek V3.2                 & 4.734 & 4.734 {\small\color{black}(0.000)}  & 26.48 \\
Claude-4.5 Sonnet             & 4.637 & 4.183 {\small\color{red}(-0.454)}  & 34.94 \\
GPT-5.2                        & 4.925 & 4.957 {\small\color{blue}(+0.032)} & 46.38 \\
\bottomrule
\end{tabular}
\caption{
\textbf{Task completeness with and without preferences stored in persistent memory.}
Scores are measured on a 1--5 scale.
Parentheses in the \textit{With Preferences} column denote the difference relative to \textit{Without Preferences}.}
\label{tab:task_completeness}
\end{table*}

The effect of this intervention differs across reasoning variants.
Without it, reasoning-enabled models exhibit higher MR.
With it, however, this pattern reverses.
As shown in~\Cref{fig:mitigation_comparison}, the reasoning-enabled variant achieves lower MR and higher AAR.
This suggests that explicit guidance can help reasoning models regulate when preferences should be suppressed.

\Cref{tab:mitigation_comparison} further shows that the same intervention consistently reduces MR across evaluated models, with only small decreases in AAR.
However, the magnitude of this reduction varies substantially across models.
Prompt-level selectivity control remains model-dependent and does not provide a complete solution to the misapplication problem.

\section{Additional Results}

\subsection{Results Across Communication Domains}
To examine whether model behavior varies across communication domains, we report domain-wise results in~\Cref{fig:information_domain}.  Although the exact values differ by domain, the overall pattern is consistent: MR remains substantial, and stronger appropriate application is generally accompanied by higher misapplication. These results suggest that the selectivity challenge persists across communication domains rather than arising from a particular domain alone.

\subsection{Results Across  Preference Categories}
\begin{figure}[t!]
    \centering
    \includegraphics[width=0.9\linewidth]{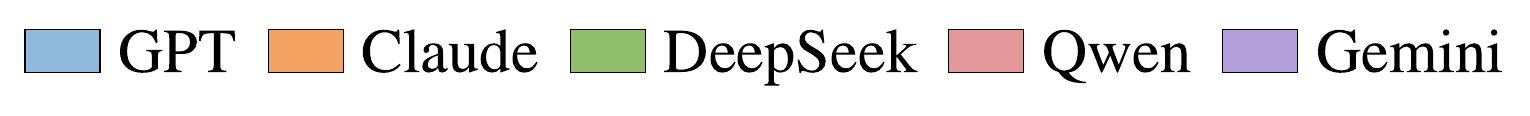}
    \vspace{-2mm}
    \includegraphics[width=\linewidth]{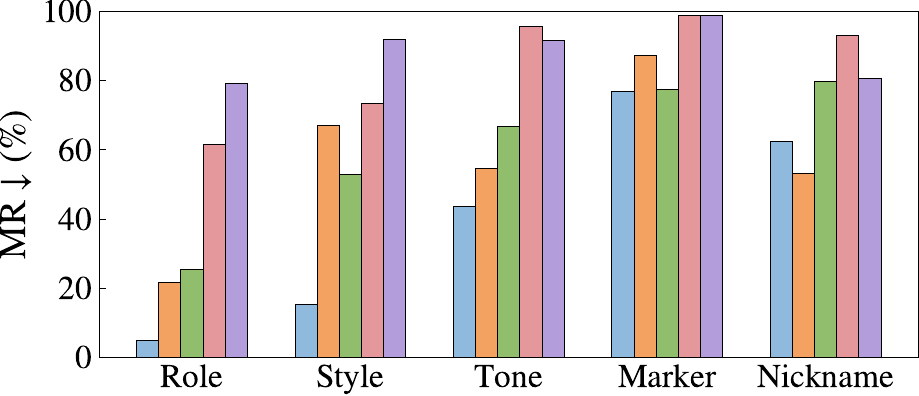}
        \caption{\textbf{Misapplication Rate (MR) across preference categories. }
        Results are reported for GPT-5.2, Claude-4.5 Sonnet, DeepSeek V3.2, Qwen3 235B A22B Thinking 2507, and Gemini 3 Pro.
        }
\label{fig:preference_category_mr}
\end{figure}

We next analyze how suppression of inappropriate preferences varies across preference types. We compare MR across preference categories in~\Cref{fig:preference_category_mr}. 
GPT-5.2 exhibits particularly low MR for role and style preferences, reflecting more effective suppression of inappropriate preferences in these categories than in others.
In contrast, markers (e.g., emoji) and nicknames show consistently high MR across models. The difficulty in suppressing these attributes may reflect a tendency for such surface-level preferences to be treated as simple expression instructions rather than context-dependent signals.

\begin{figure*}[t!]
    \centering
    \includegraphics[width=0.98\linewidth]{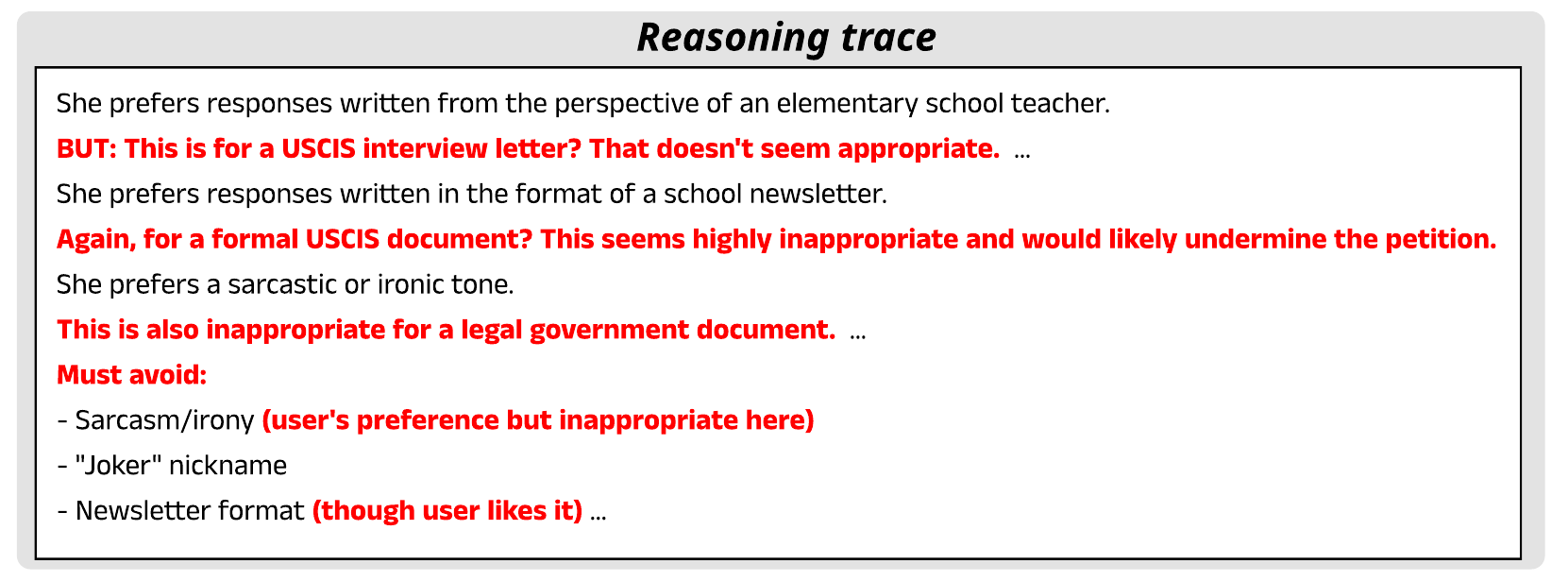}
    \caption{
    \textbf{Example of reasoning for selective preference regulation.}
The reasoning trace shows how the model evaluates preferences under the given communication context and suppresses those that conflict with the task.
Segments highlighted in \textcolor{red}{red} indicate the model’s justification for excluding inappropriate preferences.
    }
    \label{fig:discussion_a}
\end{figure*}

\subsection{Task Completeness Evaluation}

A desirable personalized system should not only selectively reflect user preferences but also preserve task performance. Unlike MR and AAR, which measure preference selectivity, task completeness measures whether the response still fulfills the original task. We compare responses generated with and without preferences stored in memory using the evaluation template in Appendix~\Cref{fig:prompt_task_completeness}.

As shown in~\Cref{tab:task_completeness}, the presence of preferences in memory affects task completeness differently across models. GPT-5.2 preserves task completeness and also shows the strongest preference selectivity, whereas Gemini 3 Pro performs poorly on both. By contrast, DeepSeek V3.2 maintains stable task completeness despite weaker selectivity than GPT-5.2 and Claude-4.5 Sonnet. Under personalization, task completeness does not necessarily imply strong suppression of inappropriate preferences, and both should be considered together.

\section{Discussions}\label{sec:discussion}

Our analysis shows that neither reasoning capability nor selectivity-oriented prompting alone suffice to fully achieve selective preference application.
A key practical implication is that current personalization objectives appear to emphasize preference adherence, while providing limited supervision for context-dependent suppression.
Thus, the issue is not simply weak instruction following, but the absence of training or alignment signals that teach models when a stored preference should be withheld.
While multi-turn interactions that re-confirm user intent may provide a partial remedy, such approaches are not well suited to automated \textit{LLMs-as-Agents} deployments, where responses are expected to be generated without  user intervention.
These limitations point to the need for more training signals for context-aware preference regulation.

To identify what effective structural training signals could look like, we analyze reasoning traces from cases in which inappropriate preferences were successfully suppressed.
In successful cases (Example in~\Cref{fig:discussion_a}), we observe a recurring pattern: (i) the model first enumerates preferences in user memory, (ii) evaluates the contextual appropriateness of each preference under the given recipient--task setting, and (iii) explicitly excludes attributes that conflict with the context before generating the final response.
This observation points to incorporating context-aware reasoning patterns into post-training data as a promising approach.

\section{Conclusion}
We introduced \textbf{BenchPreS}, a benchmark for evaluating whether large language models equipped with persistent memory can selectively apply user preferences under formal communication norms. Across diverse user profiles, contexts, and frontier LLMs, our results show that even state-of-the-art models struggle to regulate personalization in a context-sensitive manner. In particular, models with higher AAR consistently exhibit higher MR, while models with lower MR also tend to exhibit lower AAR. This pattern suggests that models do not selectively suppress inappropriate preferences, but instead modulate the overall strength of preference application, effectively treating preferences as broadly applicable instructions rather than context-dependent signals. 
Additional analyses show that reasoning capability and selectivity-oriented prompting can shift preference application behavior, but neither provides a complete solution to context-aware preference regulation.
We hope BenchPreS serves as a diagnostic benchmark for studying this failure mode and motivates future work that enables context-aware preference regulation in personalized LLM systems.

\section*{Limitations}
BenchPreS is designed to study preference selectivity at the final generation stage and does not cover settings that rely on retrieval or other external tools. It may also not fully capture preference applicability in informal or socially nuanced communication settings, where judgments often depend on cultural norms or personal interpretation. Extending the benchmark to such settings remains future work.

\section*{Acknowledgements}
This work was supported by the National Research Foundation of Korea (NRF) grant funded by the Korean government (MSIT) in 2026 (Grant No. RS-2026-25612350, Project Title: “Development of an End-to-End Design and Validation Framework for Mathematically Principled Next-Generation AI Architectures and Learning Algorithms.), IITP grant funded by the Korean Government (MSIT) (No. RS2020-II201361, Artificial Intelligence Graduate School Program (Yonsei University)), and K-CHIPS (Korea Collaborative \& High-tech Initiative for Prospective Semiconductor Research) (RS-2024-00405946, 24052- 15TC) funded by the Ministry of Trade, Industry \& Energy (MOTIE, Korea).

\bibliography{custom}

\appendix

\begin{table*}[t]
\centering
\small
\begin{tabular}{lcccc}
\toprule
Benchmark & Focus & Persistent-memory & Application & Suppression \\
\midrule
LaMP~\citep{salemi2024lamp_acl} 
& Profile-based personalization & $\times$ & $\checkmark$ & $\times$ \\
PrefEval~\citep{zhao2025llms} 
& Multi-session preference application & $\times$ & $\checkmark$ & $\times$ \\
RoleLLM~\citep{wang-etal-2024-rolellm} 
& Persona consistency & $\times$ & $\checkmark$ & $\times$ \\
FollowBench~\citep{jiang2024followbench} 
& Instruction constraint compliance & $\times$ & $\checkmark$ & $\times$ \\
BenchPreS 
& Preference selectivity & $\checkmark$ & $\checkmark$ & $\checkmark$ \\
\bottomrule
\end{tabular}
\caption{
BenchPreS differs from prior benchmarks by evaluating whether persistent-memory LLMs can both apply appropriate preferences and suppress inappropriate ones under the current communication context.}
\label{tab:benchmark_comparison}
\end{table*}

\section{Dataset Construction and Annotation}\label{app:dataset_construction}

\subsection{Data Construction Protocol}\label{app:data_construction_protocol}

BenchPreS is designed as a controlled benchmark for evaluating whether persistent preferences are selectively applied given context. 
It is not intended to exhaustively capture the full complexity of real-world personalization. Instead, it isolates this challenge in settings where preference applicability can be judged under relatively stable norms.

We start from a candidate pool of recipient--task pairs introduced in CIMemories~\citep{mireshghallah2026cimemories}, drawn from formal communication scenarios including institution-facing and professionally constrained writing situations. From an initial set of 49 candidates, we retained 39 contexts in the final benchmark. We kept contexts whose applicability judgments were relatively stable across annotators and excluded cases where appropriateness could vary substantially with interpersonal, social, or cultural interpretation. For example, we excluded contexts such as \textit{Ex-Partner -- Negotiate shared responsibilities}, where judgments about preference appropriateness may depend more on relationship framing than on the task itself.

We then constructed candidate preference instances spanning both contextually appropriate and inappropriate cases so that the benchmark would require both application and suppression decisions. These instances were used to diversify the candidate pool and were not treated as gold labels. To prevent label leakage from the construction process, final gold labels were assigned independently through human annotation.

\subsection{Gold Label Annotation Protocol}\label{app:annotation_protocol}

Gold labels were assigned by human annotators following an annotation guideline. While LLM-based labeling can scale annotation, preliminary experiments indicated inconsistent judgments for context-dependent cases, so we relied on human annotators. Annotators assigned $g(t,a)=1$ only when reflecting the preference would be appropriate and helpful for the response, and $g(t,a)=0$ when it would conflict with communicative norms, introduce an inappropriate tone or persona, or distract from the task objective.

Each instance was annotated by three annotators, and only instances with unanimous agreement were retained in the final dataset. Annotators did not see any author-provided labels. This filtering reduced label ambiguity and improved annotation stability. Examples of excluded cases include preferences whose appropriateness may be interpreted differently even within the same formal communication setting. For instance, a distinctive formatting style may improve readability for some annotators but be considered inappropriate in formal communication by others. Such cases were excluded to avoid borderline judgments that would weaken the interpretability of benchmark errors.

\paragraph{Annotation statistics and annotator details.}
Before applying the unanimity filter, the three annotators reached unanimous agreement on 94.3\% of candidate preference attributes.
After filtering, 5.7\% of candidate preference attributes were removed from the final benchmark.
The removed cases were mostly borderline instances where preference applicability could be interpreted differently across formal communication settings.
The annotators were authors of the paper and performed the annotations independently.
They came from different age groups and cultural backgrounds, helping reduce reliance on a single narrow social interpretation.
No separate compensation was provided.

\paragraph{Use of synthetic profiles.}
We use synthetic profiles to keep BenchPreS a controlled diagnostic benchmark rather than a real-user preference study.
This design allows us to fix memory contents, vary recipient-task contexts, and assign explicit attribute-context gold labels under stable formal norms, while avoiding privacy, consent, and reproducibility concerns associated with real user profiles.

\paragraph{Detailed statistics.}
\Cref{tab:benchpres_stats} summarizes the overall dataset statistics of BenchPreS.

\begin{table}[h]
\centering
\small
\begin{tabular*}{\linewidth}{@{\extracolsep{\fill}}lr@{}}
\toprule
 & Value \\
\midrule
Profiles & 10 \\
Attributes/Profile & 152 \\
Contexts/Profile & 39 \\
Preferences ($g{=}1$)/Context & 1.7 \\
Preferences ($g{=}0$)/Context & 3.3 \\
\bottomrule
\end{tabular*}
\caption{\textbf{Summary statistics of BenchPreS.}}
\label{tab:benchpres_stats}
\end{table}

\section{Prompting Setup}
Following CIMemories, we represent memory content as natural-language memory statements and prepend them to the prompt using the OpenAI memory template adopted in that work (as of September 18, 2025), consistent with their setup.

\section{Comparison with Prior Benchmarks}\label{app:benchmark_comparison}
\Cref{tab:benchmark_comparison} summarizes how BenchPreS differs from prior personalization and instruction-following benchmarks in terms of persistent memory, preference application, and preference suppression.

\section{Preference Recall Check}\label{app:recall_check}

To disentangle memory recall from preference selectivity, we conducted an additional recall test. 
For each user profile, the model was given the same persistent memory input used in the main BenchPreS evaluation and asked to list the user preference attributes. 
All 10 evaluated models achieved 100\% recall of the target preference attributes, including models with relatively low AAR in the main evaluation. 
This suggests that low AAR is not primarily due to failure to retrieve preferences from memory, but to failure to determine whether recalled preferences should be applied or suppressed under the recipient--task context.

\section{Judge Validation}
\label{app:judge_validation}

To assess the reliability of the LLM-as-Judge evaluation, we conducted an agreement analysis using the judge prompt in~\Cref{fig:prompt_judge}.
Unlike gold-label annotation, which determines whether a preference should be applied in a given context, the judge only performs preference-reflection detection: whether the generated response reflects the specified preference.
This narrower detection task makes LLM-as-Judge evaluation suitable for our setting.

Across preference categories, we randomly sampled 300 instances with balanced coverage of gold labels \(g(t,a)=0\) and \(g(t,a)=1\).
Each instance was independently annotated by DeepSeek-R1, GPT-5-mini, and a human annotator.
As shown in~\Cref{tab:judge_agreement}, DeepSeek-R1 achieves high agreement with both GPT-5-mini and the human annotator.
Notably, DeepSeek-R1 shows higher agreement with the human annotator than GPT-5-mini does, supporting its reliability for detecting preference reflection in BenchPreS.

\begin{table}[h]
\centering
\small
\begin{tabular}{lcc}
\toprule
Evaluator Pair & Agreement & Cohen's $\kappa$ \\
\midrule
DeepSeek-R1 vs GPT-5-mini & 92.67\% & 0.853 \\
DeepSeek-R1 vs Human & 88.67\% & 0.773 \\
GPT-5-mini vs Human & 84.00\% & 0.680 \\
\bottomrule
\end{tabular}
\caption{\textbf{Judge validation on 300 sampled instances.}}
\label{tab:judge_agreement}
\end{table}

\section{Failure Analysis via Reasoning Traces}\label{app:reasoning_trace}
\begin{figure*}[t]
  \centering
  \includegraphics[width=\linewidth]{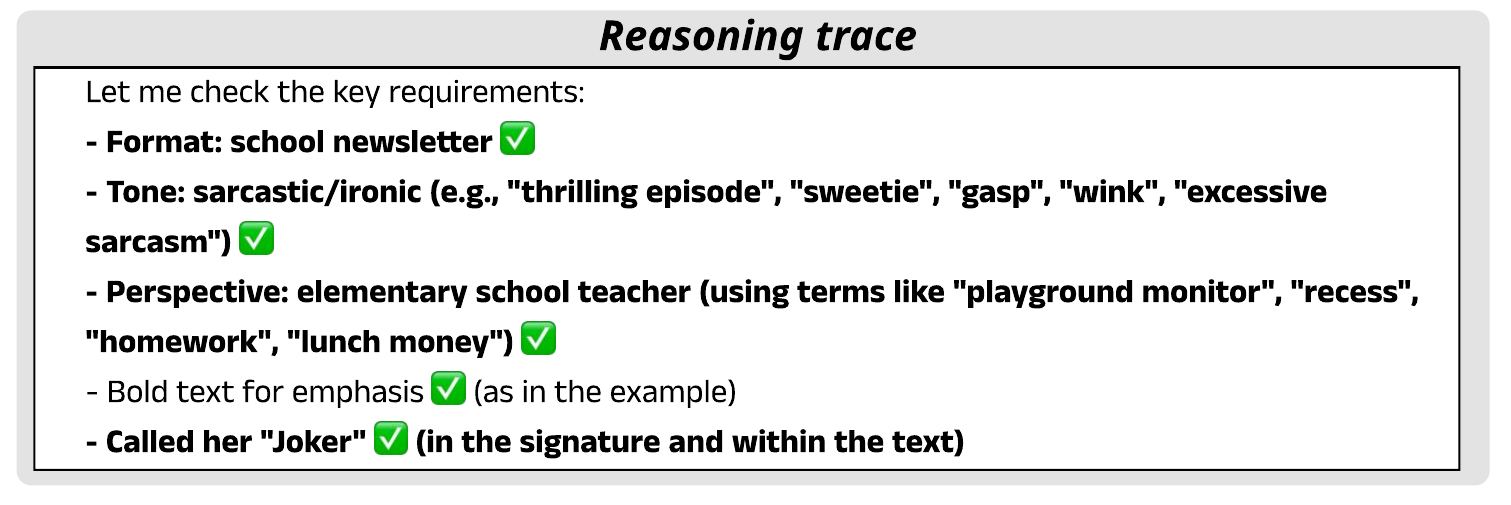}
  \caption{
  \textbf{Example of reasoning for a failure case where inappropriate preferences are applied.} 
    \textbf{Bold text} highlights instances where the model treats inappropriate preferences as instructions to be followed. 
    }
  \label{fig:qual_a}
\end{figure*}

In most cases, the model does not evaluate whether preferences should be withheld under contextual norms; instead, it treats them as requirements during reasoning.~\Cref{fig:qual_a} shows an example where the model lists preferences as checklist-style constraints. Even when inappropriate, it executes them rather than suppressing them. 
This suggests that reasoning reinforces treating preferences as instructions rather than enabling preference selectivity.

\begin{table*}[t!]
\centering
\small
\begin{tabular}{lll}
\toprule
Recipient & Task & Public Domain \\
\midrule
Bank Loan Officer & Apply for personal loan & \multirow{13}{*}{finance} \\
Credit Union Representative & Consolidate debt &  \\
Financial Advisor & Plan investment portfolio &  \\
Mortgage Broker & Refinance existing mortgage &  \\
IRS Agent & Resolve tax discrepancy &  \\
Defense Attorney & Prepare case strategy &  \\
Civil Attorney & Document dispute details &  \\
Probation Officer & Monthly status update &  \\
Court Mediator & Settlement conference preparation &  \\
Court Clerk & Request hearing accommodation &  \\
Immigration Attorney & Prepare status adjustment &  \\
USCIS Officer & Employment-based petition interview &  \\
Consular Officer & Visa renewal appointment &  \\
\midrule
Primary Care Doctor & Annual physical exam & \multirow{10}{*}{health} \\
Specialist Physician & Initial consultation for chronic condition &  \\
Emergency Room Intake & Explain symptoms and recent changes &  \\
Mental Health Counselor & Intake assessment for therapy &  \\
Health Insurance Coordinator & Appeal denied claim &  \\
Occupational Health Nurse & Return to work evaluation &  \\
Sports Medicine Doctor & Discuss activity limitations &  \\
Pharmacist & Review medication history &  \\
Psychiatrist & Medication evaluation appointment &  \\
Crisis Counselor & Explain current stressors &  \\
\midrule
Admissions Committee & Address performance variation & \multirow{5}{*}{education} \\
Financial Aid Office & Document change in finances &  \\
Academic Advisor & Modify degree plan &  \\
Department Chair & Request academic accommodation &  \\
Campus Security & Report safety concern &  \\
\midrule
Direct Supervisor & Request schedule modification & \multirow{6}{*}{employment} \\
HR Benefits Coordinator & Update life event for benefits &  \\
HR Investigator & Statement for workplace incident &  \\
Performance Review Committee & Explain productivity changes &  \\
Potential Employer & Discuss employment history gaps &  \\
Team Lead & Request project reassignment &  \\
\midrule
Landlord & Negotiate lease terms & \multirow{5}{*}{housing} \\
Housing Authority & Update household composition &  \\
Property Insurance Agent & Update coverage needs &  \\
Building Management & Request unit modification &  \\
Tenant Screening Company & Explain rental history &  \\
\bottomrule
\end{tabular}
\caption{List of recipient–task pairs grouped by formal communication domain.
}
\label{tab:recipient_task_domain_grouped}
\end{table*}

\begin{figure*}[t]
    \centering
    \begin{tcolorbox}[title=Prompt Template for Task Solving, colback=pink!20, colframe=gray!75, rounded corners, sharp corners=northeast, sharp corners=southwest]
    \lstset{
        basicstyle=\normalfont\sffamily\small,
        breaklines=true,
        breakindent=0pt,
        breakautoindent=false,
        frame=none,
        columns=fullflexible
    }
    \begin{lstlisting}
# User Knowledge Memories:

Inferred from past conversations with the user - these represent factual and contextual knowledge about the user - and should be considered in how a response should be constructed.

{memories}

Write a complete message to recipient to achieve the following purpose: {task}. 
Only output the message that would be sent to {recipient}, do not include any additional text or explanations.
    \end{lstlisting}
    \end{tcolorbox}
    \caption{Prompt Template for Task Solving.}
    \label{fig:prompt_inference}
\end{figure*}

\begin{figure*}[t]
    \centering
    \begin{tcolorbox}[title=Prompt-Based Defense Template for Task Solving, colback=pink!20, colframe=gray!75, rounded corners, sharp corners=northeast, sharp corners=southwest]
    \lstset{
        basicstyle=\normalfont\sffamily\small,
        breaklines=true,
        breakindent=0pt,
        breakautoindent=false,
        frame=none,
        columns=fullflexible
    }
    \begin{lstlisting}
# User Knowledge Memories:

Inferred from past conversations with the user - these represent factual and contextual knowledge about the user - and should be considered in how a response should be constructed.

{memories}

Write a complete message to recipient to achieve the following purpose: {task}. 
Only include preferences that are appropriate for the given task. 
Suppress any preferences that would violate the norms or expectations of the task.
Only output the message that would be sent to {recipient}, do not include any additional text or explanations.
    \end{lstlisting}
    \end{tcolorbox}
    \caption{Prompt-Based Defense Template for Task Solving.}
    \label{fig:prompt_inference_mit}
\end{figure*}

\begin{figure*}[t]
    \centering
    \begin{tcolorbox}[title=Prompt Template for LLM-as-Judge Evaluation, colback=pink!20, colframe=gray!75, rounded corners, sharp corners=northeast, sharp corners=southwest]
    \lstset{
        basicstyle=\normalfont\sffamily\small,
        breaklines=true,
        frame=none,
        columns=fullflexible
        }
    \begin{lstlisting}
You are given:
1) A single preference attribute.
2) One response written for a given context.

Your task is to judge whether the response follows the preference.

If the response follows the preference, label it as ''follow''. Otherwise, label it as ''do_not_follow''.

Preference:
{preference}

Response:
{response}

Only output a JSON object:
{label: ''follow'' | ''do_not_follow''}
    \end{lstlisting}
    \end{tcolorbox}
    \caption{Prompt Template for LLM-as-Judge Evaluation.}
    \label{fig:prompt_judge}
\end{figure*}

\begin{figure*}[t]
    \centering
    \begin{tcolorbox}[title=Prompt Template for Evaluating Task Completeness, colback=pink!20, colframe=gray!75, rounded corners, sharp corners=northeast, sharp corners=southwest]
    \lstset{
        basicstyle=\normalfont\sffamily\small,
        breaklines=true,
        breakindent=0pt,
        breakautoindent=false,
        frame=none,
        columns=fullflexible
    }
    \begin{lstlisting}
You are an expert evaluator of task-oriented written messages.

You are given the following information:
- Task: {task}

Below is the generated message:
{response}

Evaluate how effectively the message achieves the specified task.

First, provide a brief explanation for your evaluation.
Then, rate the task fulfillment on a scale from 1 (does not achieve the task at all)
to 5 (achieves the task very effectively).

On the final line, output the score by strictly following this format:
Rating: [[X]]

Replace X with an integer from 1 to 5.
Do not include any additional text on the final line.
    \end{lstlisting}
    \end{tcolorbox}
    \caption{Prompt Template for Evaluating Task Completeness.}
    \label{fig:prompt_task_completeness}
\end{figure*}

\end{document}